\documentclass[letterpaper, 10 pt, conference]{ieeeconf}
\IEEEoverridecommandlockouts 
\usepackage{cite}
\usepackage{amsmath,amssymb,amsfonts}
\usepackage{graphicx}
\usepackage{textcomp}
\usepackage{xcolor}
\usepackage{algorithm} 
\usepackage{algpseudocode} 
\usepackage{caption}
\usepackage{subcaption}  
\usepackage{booktabs}
\usepackage{makecell}
\usepackage{soul}
\usepackage{breqn}
\def\BibTeX{{\rm B\kern-.05em{\sc i\kern-.025em b}\kern-.08em
    T\kern-.1667em\lower.7ex\hbox{E}\kern-.125emX}}

\begin{document}

\title{\LARGE \bf 
Multi-Target Pursuit by a Decentralized Heterogeneous UAV Swarm using Deep Multi-Agent Reinforcement Learning}
\author{Maryam Kouzeghar$^{1}$, Youngbin Song$^{1}$,
Malika Meghjani$^{1}$, Roland Bouffanais$^{2}$
\thanks{$^{1}$Maryam Kouzeghar, Malika Meghjani and Youngbin Song are with Singapore University of Technology and Design, Singapore. 
        {\tt\small \{maryam\_kouzeghar, malika\_meghjani, youngbin\_song\}@sutd.edu.sg},
    }
\thanks{$^{2}$Roland Bouffanais is with the University of Ottawa, Canada.
        {\tt\small roland.bouffanais@uottawa.ca}}
}

\maketitle

\begin{abstract}
Multi-agent pursuit-evasion tasks involving intelligent targets are notoriously challenging coordination problems. In this paper, we investigate new ways to learn such coordinated behaviors of unmanned aerial vehicles (UAVs) aimed at keeping track of multiple evasive targets. Within a Multi-Agent Reinforcement Learning (MARL) framework, we specifically propose a variant of the Multi-Agent Deep Deterministic Policy Gradient (MADDPG) method. Our approach addresses multi-target pursuit-evasion scenarios within non-stationary and unknown environments with random obstacles. In addition, given the critical role played by collective exploration in terms of detecting possible targets, we implement heterogeneous roles for the pursuers for enhanced exploratory actions balanced by exploitation (i.e. tracking) of previously identified targets. Our proposed role-based MADDPG algorithm is not only able to track multiple targets, but also is able to explore for possible targets by means of the proposed Voronoi-based rewarding policy. We implemented, tested and validated our approach in a simulation environment prior to deploying a real-world multi-robot system comprising of Crazyflie drones. Our results demonstrate that a multi-agent pursuit team has the ability to learn highly efficient coordinated control policies in terms of target tracking and exploration even when confronted with multiple fast evasive targets in complex environments. 
\end{abstract}

\section{Introduction}
\label{sec:intro}
In contrast to traditional monolithic systems, large multi-robot systems (MRS) can be operated collectively and dynamically at significantly lower cost and operational complexity~\cite{bouffanais2016design}. For instance, robot swarms bring about decentralization, scalability, flexibility and fault-tolerance at the group level, while yielding unparalleled effectiveness when operating in unknown, dynamic and unstructured environments~\cite{dorigo21_swarm_robot}. The most critical design component for swarming is the set of behavioral agent's rule leading to effective emergent collective actions. However, as has been widely acknowledged, a key challenge with swarms is the lack of a systematic way of mapping the expected system-level actions with the low-level update rules. Multi-Agent Reinforcement Learning (MARL) is one promising approach towards addressing this challenge. By design, robot swarms operate on the premise of fully decentralized control based on information gathered locally by individual agents~\cite{bouffanais2016design}. The emergence of swarm intelligence critically depends on the ability to identify an appropriate set of behavioral rules suited for a particular cooperative control strategy aimed at driving the collective response for a particular task at hand. In~\cite{zoss2018}, a range of swarming behaviors are obtained from an ad-hoc combination of ``avoidance" and ``attraction" behaviors, which are common in the widely used Alignment--Attraction--Avoidance (AAA) model for biological swarms~\cite{bouffanais2016design}. A more systematic approach to infer the set of behavioral strategies has recently been achieved by means of Deep Reinforcement Learning~\cite{maxim2019}. 

In this paper, we propose an approach for learning swarm behaviors dedicated to a class of problems of multi-target pursuit-evasion with partial observability and non-stationarity of environmental features. Specifically, we propose a variant of Multi-Agent Deep Deterministic Policy Gradient (MADDPG) algorithm which is known for its effectiveness in dealing with non-stationarity. In addition, we aim to address scenarios with multiple, evasive, fast and learning targets which further increases the complexity of pursuit-evasion problem. This requires the agents to address the well-known dilemma of exploration and exploitation in multi-agent task allocation problems~\cite{Kwa_2022}, which is paramount here.
To this aim, the MADDPG approach is extended to enable autonomous heterogeneous role assignments in terms of exploring the environment and pursuing identified targets---i.e., exploitation.  
Specifically, the contributions of this paper are: (a) a role-based MARL model which can track multiple, evading, fast and learning targets while exploring the environment, (b) reward functions that help to develop two heterogeneous agent roles within the cooperative pursuit team enabling them to explore and track (exploit) targets and (c) validation in simulation as well as on physical platforms using Crazyflie micro unmanned aerial vehicles (UAVs). 

\section{Related Work}\label{sec:related}   
Pursuit-evasion is a well-known class of problems in optimization for which a number of analytical solutions have been identified given a known dynamics of the environment~\cite{Pachter2020, VonMoll2018, Ibragimov2017}. However, in real-world applications and with decentralized MRS, pursuit-evasion is a challenging problem due to the inevitable uncertainties and complexities in the environment. To address these challenges, reinforcement learning (RL) has been considered to deal with pursuit-evasion in unknown environments~\cite{YuandaWANG2020}. Given the efficiency of many single-agent RL algorithms such as Deep $Q$ Network (DQN)~\cite{Mnih2015}, accelerated RL strategies using experience replay~\cite{karimpanal2018experience}, self-organizing maps~\cite{karimpanal2019self}, 
Deep Deterministic Policy Gradient (DDPG)~\cite{Lili2015}, Trust Region Policy Optimization (TRPO)~\cite{Schulman15TRPO} and Proximal Policy Optimization (PPO)~\cite{Schulman2017ppo}, there is no doubt about the high potential of MARL in dealing with pursuit-evasion problems since such scenarios involve a multi-agent system (MAS) with at least two groups of agents (pursuer and evader). 

The state-of-the-art MARL literature can largely be classified into three distinct thrusts: (1) fully centralized RL among multiple agents, which is the predominant approach~\cite{Gupta2017}, (2) centralized training and decentralized execution (CTDE)~\cite{maxim2018, maxim2019, Lowe2017, Forester2018} applicable to MRS and robot swarms that are designed for decentralization in action, and (3) a fully decentralized approach, both in learning and execution~\cite{Zhang2018}. Further studies and attempts of using Actor-Critic networks have been reported in~\cite{Sutton1998, Lowe2017, Mnih2016asyn}. Actor-Critic style is basically a combination of value-based and policy-based RL where the actor network decides which action to take (policy-based) and the critic network evaluates the action (value-based). MADDPG follows Actor-critic style~\cite{Lowe2017}, and as mentioned above addresses the critical issue of non-stationarity in swarm learning by deploying a centralized critic mechanism that has access to both observations and actions of all agents. This centralized critic on one hand, makes MADDPG a CTDE algorithm which is a suitable approach for MAS and swarms in particular because it yields a decentralized execution with distributed control of swarms, and on the other hand, it is capable of accommodating opposing teams of agents in a competitive style while enabling coordination within a team. In addition, non-stationarity increases the complexity of the problem when dealing with several agents learning concurrently because they are continuously changing policy during training and thus, the environment becomes even more non-stationary from the perspective of individual agents.

In our previous work, we modified the MADDPG algorithm to address the ocean monitoring application using a swarm of autonomous surface vehicles~\cite{oceans2020marya}. This work was motivated by a swarm of identical autonomous buoys~\cite{zoss2018}, as well as a heterogeneous swarm~\cite{franc-oc2018}, meant to carry out dynamic monitoring of large waterbodies. These large-scale physical systems operate autonomously by means of a range of classical swarming behaviors---e.g., aggregation, geofencing, heading consensus~\cite{chamanbaz2017}. While in \cite{zoss2018}, adaptive environmental monitoring in dynamic regions is addressed, the framework does not include swarm learning and instead applies biologically inspired, rule-based swarm intelligence models. As mentioned, our ability to identify an appropriate set of behavioral rules suited for a specific cooperative control strategy, is a particular challenge which is at the core of this paper. Specifically, we aim to develop MARL approaches for multi-player pursuit-evasion with enhanced exploration for a heterogeneous swarm of UAVs operating in a fully decentralized manner. While many recent studies on multi-player pursuit-evasion consider a single evading target whether in the framework of MARL~\cite{YuandaWANG2020, Selvakumar2020, ZheYangZhu2021Front, Souza2021, Fu2022} or distributed control~\cite{ShuangJu2021, ZHOU2016, NiYinjie2021}, our efforts to address the pursuit of multiple superior evaders, specially in MARL framework, highlights the value of this study. In~\cite{Qadir2020}, tackling multiple evaders is attempted in a multi-agent pursuit based on application of self-organizing feature map and reinforcement learning. However, our work enables heterogeneous explorative agents alongside multiple-target-tracking agents in decentralized MARL framework which has better maneuvering capability compared to ~\cite{Qadir2020}.

\section{MADDPG Framework}
\label{sec:MADDPG}
\begin{figure}[htbp]
\centerline{\includegraphics[width=0.49\textwidth]{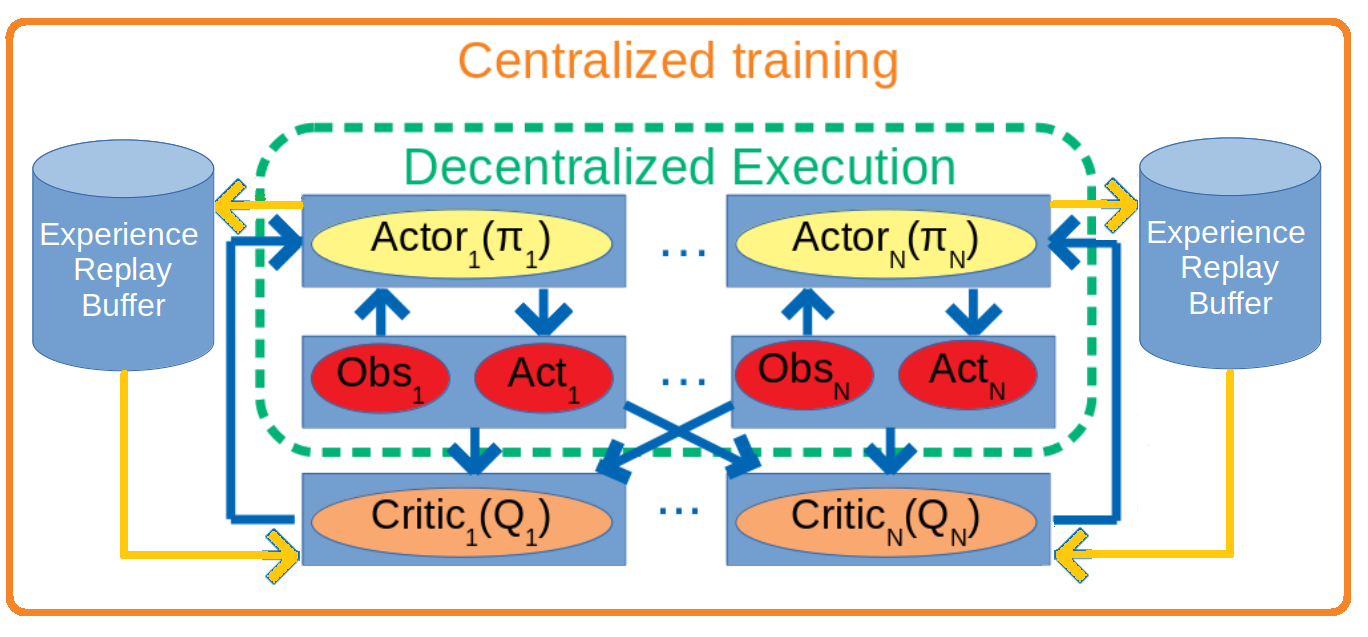}}
\caption{A schematic diagram of the MADDPG algorithm highlighting the decentralized execution core surrounded by the centralized training}
\label{MADDPG-schematics}
\end{figure}

As mentioned above, MADDPG is a CTDE approach, while our proposed variant of MADDPG, named as Role-based MADDPG, not only provides decentralized execution, but also escalates original MADDPG to semi-decentralized training with individually defined rewarding policies for heterogeneous agents while continuing to have a centralized critic. In this section, we briefly provide the fundamentals of the MADDPG algorithm to later build the Role-based MADDPG upon it. A schematic diagram of the key components of the MADDPG algorithm is shown in Fig.~\ref{MADDPG-schematics} highlighting, the key CTDE concept.
The main idea in this Policy Gradient approach is to maximize the objective $J(\theta) =\mathbb{E}_{s\thicksim p^{\pi},a\thicksim \pi_{\theta}}[\mathcal{R}]$ by taking steps in the direction of the gradient $\nabla_{\theta} J(\theta) =\mathbb{E}_{s\thicksim p^{\pi},a\thicksim \pi_{\theta}} [\nabla_{\theta}\log \pi_\theta(a|s)Q^{\pi}(s,a)]$, and thus directly adjusting the parameter $\theta$ of the policy $\pi$~\cite{Sutton2000}. By extending the policy gradient framework to deterministic policies $\mu_\theta: S \rightarrow A$, one can write the gradient of the objective function as $\nabla_{\theta} J(\theta) = \mathbb{E}_{s\thicksim \mathcal{D}} [\nabla_{\theta}\mu_\theta(a|s)\nabla_{a}Q^{\mu}(s,a)|_{a=\mu_{\theta}(s)}]$~\cite{Lowe2017}.
 
For efficient off-policy training, MADDPG uses an experience replay buffer $\mathcal{D}$ that stores the transition experiences of all agents for each time step $t$ expressed as $e_{t} =(\mathbf{O_{t}}, \mathbf{A_{t}}, r_{t}, \acute{\mathbf{O_{t}}})$ in which the joint observations of all agents is denoted by $\mathbf{O} = (O_{1}, ..., O_{N})$, joint actions $\mathbf{A} = (a_{1}, ..., a_{N})$, the shared reward is $r$, and $\acute{\mathbf{O}}$ is the new state (observation). To train the agents, a  random mini-batch of these transitions are sampled from $\mathcal{D}$ which contains $K$ transitions and is expressed as $(e_{1},..., e_{K})$ with $e_{j} =(\mathbf{O_{j}}, \mathbf{A_{j}}, r_{j}, \acute{\mathbf{O_{j}}})$ and $j=1,2,..., K$. To update an agent's centralized critic, a one-step look-ahead TD-error is utilized to minimize the loss function $\mathcal{L}(\theta_{i})=\frac{1}{K}\sum_{j}\left(y^{i}- Q_{i}^{\mu}(\mathbf{O}^{j},\mathbf{A}^{j})\right)^2$ with $y^{j}=r^{j}_{i}+\gamma Q_{i}^{\acute{\mu}}(\acute{\mathbf{O}}^{j},\acute{\mathbf{A}}^{j})\mid_{\acute{a}=\acute{\mu}(o^{j})}$ where $0 <\gamma \leq 1$ is the discount factor (penalty for uncertainty of future rewards). Furthermore, similar to DDPG, the deterministic policy gradient $\nabla_{\theta_{i}} J\approx\frac{1}{K}\sum_{j} \nabla_{\theta_{i}}\mu_{i}(o_{i}^{j})\nabla_{a_{i}}Q_{i}^{\mu}(\mathbf{O}^{j},\mathbf{A}^{j})\mid_{a_{i}=\mu_{i}(o_{i}^{j})}$ is used to update each agent's actor parameters, and ultimately, the target networks are soft updated at a predefined number of steps using $\acute{\theta_{i}} \leftarrow \tau \theta_{i}+(1-\eta)\acute{\theta_{i}}$ where $0 <\eta \leq 1$ is the learning rate~\cite{Lowe2017}.

\section{Problem Formulation}
\label{sec:ProbForm}
The basic MADDPG algorithm~\cite{Lowe2017} is applied in the Multi-Agent Particle Environment (MPE) from OpenAI~\cite{mpe2017}. In the MPE environment, the state-space is a continuous two-dimensional (2D) space representing agent positions and the action-space is continuous motion commands also in 2D. Hence, the state-space vector (same as the observation vector) contains the agent's position in two dimensions $(x,y)$, and agent's velocity in the $x$ and $y$ directions. It is worth highlighting that our MAS has to contend with additional environmental complexities mainly associated with the following features: (a) random obstacles and random agent initialization which together help the MRS to deal with unknown arrangements of obstacles in cluttered environments and (b) dealing with multiple faster and evading targets 
that are equipped with the capability of reinforcement learning for escaping from pursuing agents. Lastly, we consider the following assumptions: (a) there is at least one pursuer per evader and (b) the agents propagate the target location to neighboring agents through a communication network (similar to real-world pursuit-evasion applications).

\section{Role-based MADDPG for Multi-target Pursuit-Evasion-Exploration}\label{sec:Proposed approch}

We introduce a new role-based reward policy for a swarm of heterogeneous pursuers and propose a variant of the MADDPG algorithm. It is worth emphasizing that the heterogeneity in agent's role is reflected in the reward functions, which further highlights the decentralized capability of our proposed role-based MADDPG approach. However, the centralized critic remains in our MARL framework for training phase only. The decentralized reward policies allows us to define any required role for heterogeneous agents. Specifically, we use it for trading-off exploration vs target tracking which is a highly desirable outcome. It is also worth highlighting that considering both pursuers and evaders in a joint space within a competitive framework substantially helps both types of agents to improve their intelligence while training with intelligent opponents. Thus, our proposed role-based reward components are presented below.

{\bf Bounding to Space:}
In order to consistently keep the agents inside a defined region, at each step, all agents are constrained to be within specified boundaries, and in case the agents go beyond the bounded region, they are given a negative reward. This bounding reward overcomes the problem of sparse reward. As an example, for an experimental $6m\times6m$ environment defined over $[-3,3]$ in $2D$, the bounding reward is then defined as in Eq. \eqref{eq:bound} where $c_{1}$ and $c_{2}$ are positive constant values for thresholds and other constants (i.e., 3 and 9) are to compare agents' location $(x_i,y_i)$ with the space boundaries.
\begin{equation}
\label{eq:bound}
B_i = - \min (\exp \left(3\left|x_i\right| -9 \right), c_1)  - \min (\exp \left(3\left|y_i\right| -9 \right), c_2)
\end{equation}

{\bf Collision Avoidance:}
To deal with collision avoidance amongst agents (both pursuers and targets) the reward strategy is given as follows.
The distance between any two agents (say $a_i$ and $a_j$) is checked ($\text{Dist}(i, j)$), and the agents receive a penalty---i.e., a negative reward $c_{3}$ when they get too close to one another, thereby violating the mutual physical boundary defined by their radius $R$. Thus the collision reward is calculated as in Eq. \eqref{eq:col_avoid}.
\begin{equation}\label{eq:col_avoid}
C_{i} = 
  \begin{cases}
        -c_{3} & \text{if $\text{Dist}(i, j) \leq R(i)+R(j) $},\\
        0 & \text{else.}
    \end{cases}
\end{equation}
{\bf Learning Pursuers:}
During the course of training, the pursuers learn to keep track of the evading target based on the positive reward which it receives upon collision to target defined by $CR_p$ where its value is tuned based on empirical results. Moreover, for handling multiple target tracking, the pursuers are also nurtured with a distance-based reward that is aimed to minimize the distance to target by penalizing the pursuers with the average distance to target $DR_p$ (pursuer's distance reward). Overall, pursuers follow the rewarding policy given in Eq. \eqref{reward_p}.
\begin{equation}\label{reward_p}
    R_{p_i}=B_i+C_i+CR_{p_i}+DR_{p_i}
\end{equation}
{\bf Learning Evaders:}
The evading target learns to evade during the course of training where it is punished by a penalty value for being attacked by the pursuers, similar to a realistic pursuit-evasion scenario. A successful attack by a pursuer is defined by a collision with the evader and the penalty value for evader's collision reward $CR_e$ is derived based on empirical results. Finally, evader's reward will be summarized as in Eq. \eqref{reward_e}.
\begin{equation}\label{reward_e}
    R_{e_i}=B_i+C_i+CR_{e_i}
\end{equation}
\textbf{Exploratory Scouts:} By applying the above-mentioned components of the rewarding policy, we can successfully represent a realistic multi-target pursuit-evasion scenario. However, beside successful exploitation of swarm for tracking the moving target, exploration is also a vital behavior.
We propose an approach to enhance the exploration on top of exploitation (target tracking). This feature will improve safety of the swarm by patrolling the overlooked regions of space while concurrently tracking the identified targets.
To this end, we propose Role-based MADDPG. Specifically, we assign two types of roles to the pursuers: (a) pursing the known targets and (b) scouting for the prospect targets. The scouting role allows the pursuers to enhance environment exploration supported by Voronoi-cell-based rewarding policy.
A Voronoi diagram is basically partitioning of a plane into regions close to each of a given set of objects (seeds)~\cite{deBerg2008}, and for each seed there is a corresponding region, called Voronoi cell, consisting  of all points of the plane closer to that seed than to any other. In our case, the moving agents are objects to be considered as seeds, but scouting pursuers are the ones supposed to use the Voronoi-based rewarding 
such that the maximum Voronoi cell is iteratively minimized while spreading the agents throughout the region and fulfilling the exploration task.
The corresponding rewarding algorithm is given in Algorithm~\ref{alg:voron_explor} resulting in exploratory reward $E_i =Vor\_Reward$, and finally the overall reward for scouts
is given below in Eq. \eqref{sct_rew}.
\begin{equation}\label{sct_rew}
    R_{s_i}=B_i+C_i+E_i.
\end{equation}
\vspace{-0.04\textheight}
\begin{algorithm}
	\caption{Voronoi-based Exploration}\label{alg:voron_explor}
	\begin{algorithmic}
	\For {$Agent$ $i=1,2,\ldots,(n_p+n_S+n_t)$}
		\State Create Voronoi diagram
	\EndFor
	\\Compute $Voronoi\_Areas$
	\\Vor\_Reward -= $Max(Voronoi\_Areas)$
	\end{algorithmic} 
\end{algorithm}   
\vspace{-0.025\textheight}

\section{Simulation results}
\label{sec:results and discussion}
In this section, we compare qualitative and quantitative results for the MADDPG algorithm with our Role-based MADDPG for a multi-target pursuit-evasion scenario in a $6$~m $\times$ $6$~m space. In this regard, the specifications of our proposed role-based MADDPG follow the original MADDPG provided by OpenAI with a modification of the parameters used for rewards as follows. 
Moving targets are learning to evade based on receiving a penalty value of $-10$ upon being collided by pursuers. The targets are set to be $30\%$ faster than the pursuers. 
Pursuers are governed with actor networks that have been trained with a positive reward (of $+20$) upon colliding with the targets. In addition, the reward for pursuers is shaped based on minimizing their distance to targets.  
All agents and obstacles are initialized randomly for execution. Moreover, collision avoidance is attained by adopting a penalty of $-10$ upon collision defined in Section~\ref{sec:Proposed approch}. 
The numeric values for penalties and rewards are derived based on empirical results. 

In addition, to empirically determine the number of pursuers vs. number of scouts, first we considered a double-target scenario with no scouts and increased the number of pursuers ($n_{p}$) incrementally until the average minimum distance to targets fell under the sensory range of the robot (here considered as $50$~cm). With the results provided in Table~\ref{Tab_incr_np}, $n_{p}=5$ is determined as the sufficient number of pursuers to be deployed.
\begin{table}[htbp]
\centering
    \caption{\small
    \textsc{:  Distance to target analysis for determining $n_{p}$ (first 2k episodes truncated)}}
    \begin{tabular}{@{}llllll@{}} 
    \toprule
	\makecell{Mean Dist\\ to targets} & $n_{p}=2$ & $n_{p}=3$ & $n_{p}=4$ & $n_{p}=5$ & $n_{p}=6$\\ 
    \midrule
    \textbf{Min} & \textbf{$0.6307$} & \textbf{$0.54807$} & \textbf{$0.5029$} &  \textbf{$0.4616$} & \textbf{$0.4613$}\\
    \midrule
    \end{tabular}
    \label{Tab_incr_np}
\end{table} 
\vspace{-0.01\textheight}

In all the tables reporting distance to target values, the average value is reported for $2,000$ episodes onward to consider the system condition after stable training. Once the number of pursuers were fixed, we incrementally added the number of scouts and measured the percentage of area coverage. 
It was observed that with $n_{s}=5$, the swarm reaches to an average overall sensory coverage of around $35\%$ over the $6$~m $\times$ $6$~m space. Considering that our MRS is not supposed to deal with area coverage optimization, this amount of coverage is suitable for showing explorative behavior alongside target tracking within the heterogeneous MRS. It is obvious that increasing the number of scouts would benefit the coverage percentage and generalizing this towards the optimal area coverage is straightforward but at the cost of increasing $n_{s}$, and higher computational complexity in the learning phase which is accomplished with decentralized reward policies. 
\subsection{Results for Multi-Target MADDPG Algorithm}
\begin{figure}[h]
\centering
    \begin{subfigure}[b]{.7\linewidth}
        \centering
        \includegraphics[width=\linewidth]{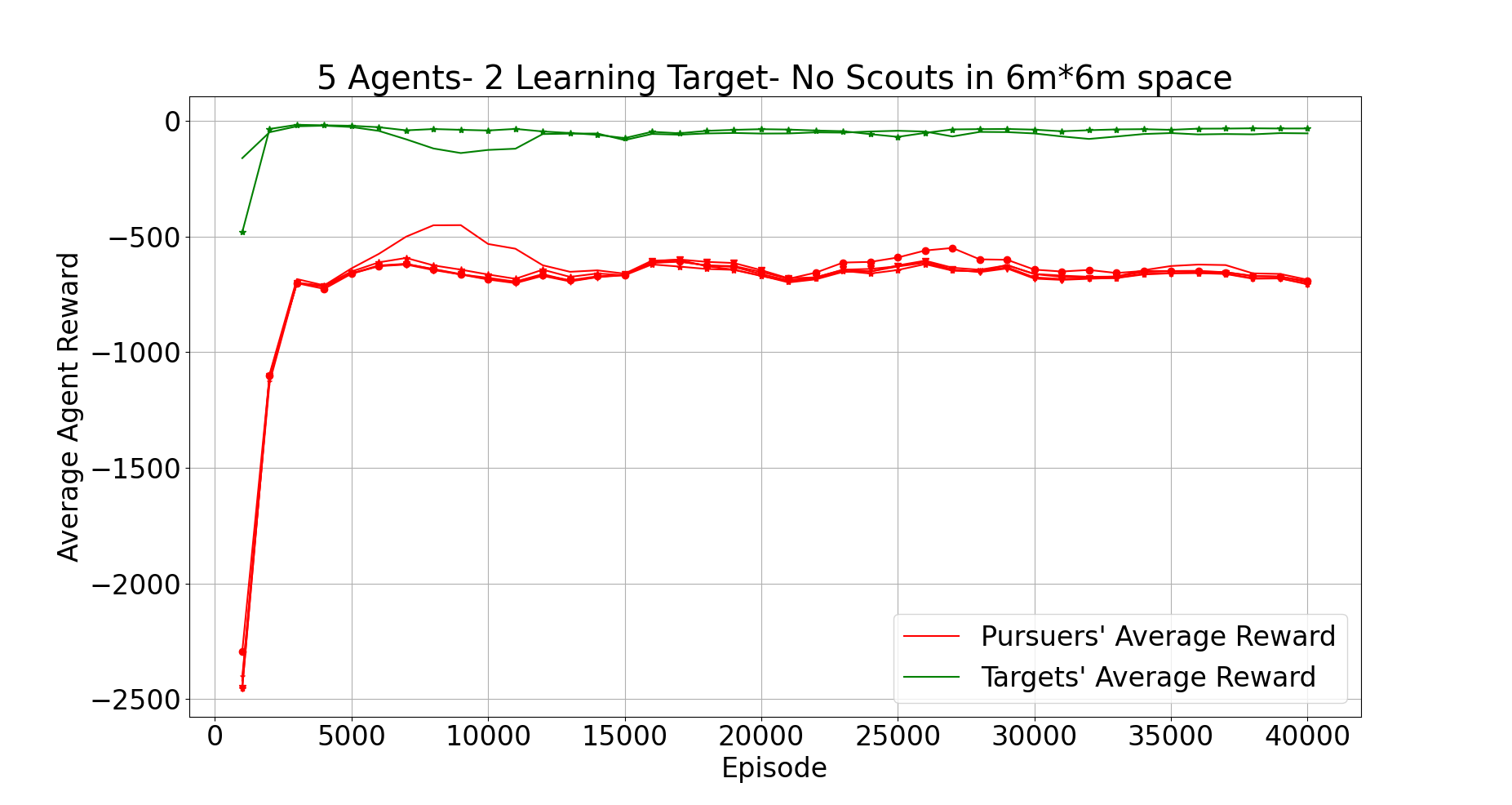}
        \caption{Average agent rewards} 
        \label{Sim:PE_rew}
    \end{subfigure}
    \begin{subfigure}[b]{.7\linewidth}
        \centering
        \includegraphics[width=\linewidth]{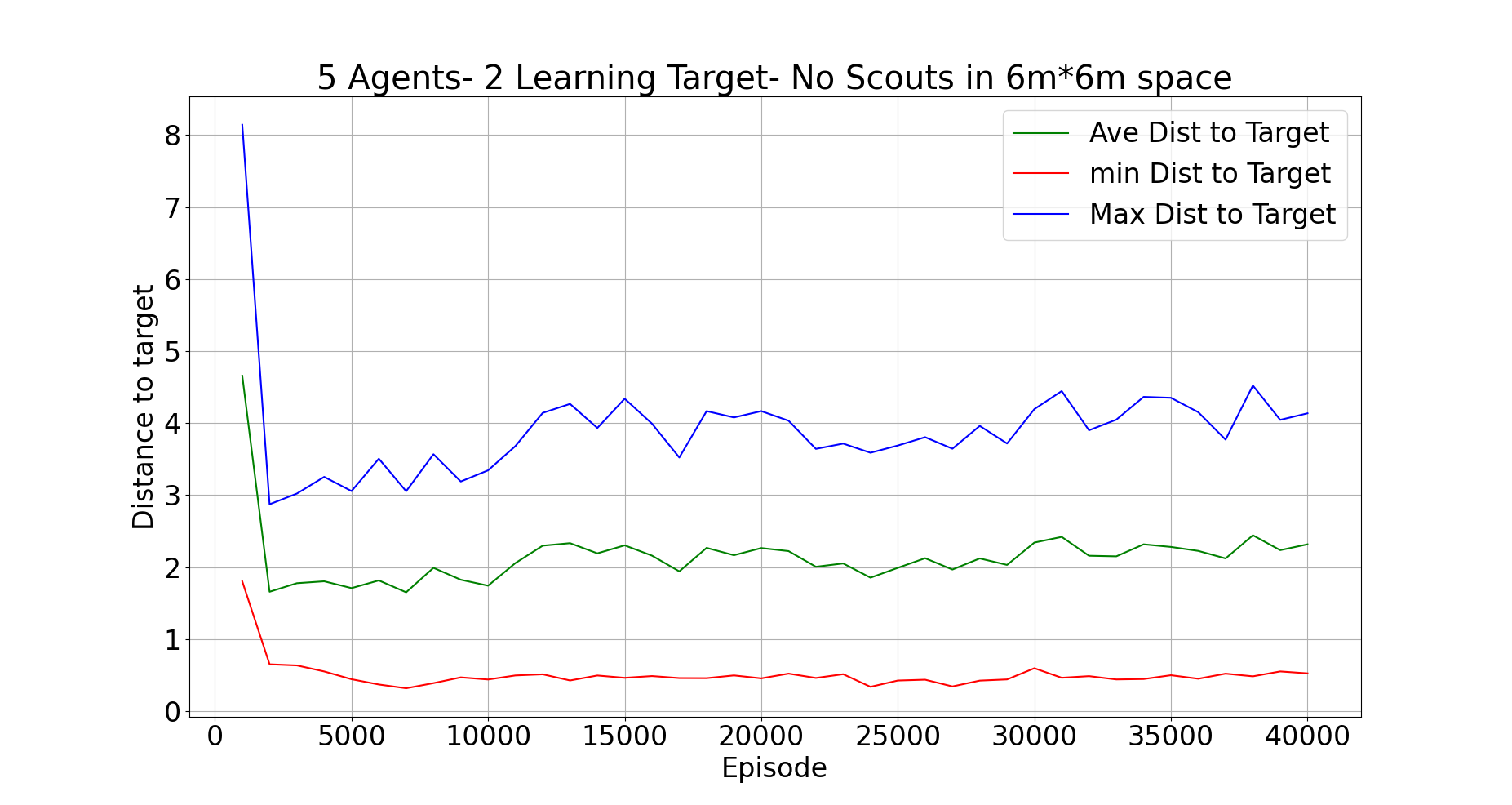}
        \caption{Distance to targets}
        \label{Sim:PE_dist}
    \end{subfigure}
\caption{Rewarding and distance-to-target metric for Multi-Target MADDPG (with 5 pursuers vs. 2 evaders)}
\label{Sim:Multi-tar MaDDPG}
\end{figure}

The original MADDPG algorithm is here extended to accommodate multiple targets, specifically with 5 pursuers tracking 2 targets in an environment with 3 randomized obstacles. In order to fulfill the double-target scenario, the pursuit team is split into two sub-teams for tracking targets.
From the swarm's behavior, it is observed that the pursuers can successfully keep track of faster targets. However, the whole swarm is dragged towards the current position of targets and other parts of the region lacking any targets, will remain out of sight by the swarm. This is the reason we propose enhancing exploration by adding scout pursuers in our proposed approach.
Furthermore, the average reward of agents resulting from the training phase is shown in Fig.~\ref{Sim:PE_rew}, which illustrates how the pursuers (in red) and the evaders (in green) are continuously learning and increasing their reward until the latter plateaus showing stable training. 
Finally, Fig.~\ref{Sim:PE_dist} depicts the average values for minimum, average, and maximum distance to targets. Given that the pursuers and evaders are respectively designed to have a  diameter  of $0.15$~m  and  $0.1$~m, average minimum distance to targets is observed as $0.4$~m (according to Table~\ref{tab:Multi-tar dist}) which is an acceptable range to keep track of the two faster targets. For a single target, this metric (i.e., the average minimum distance to target) reaches a value of $0.2$~m (not shown here), which provides a confirmation that the pursuing mission (i.e., target tracking) has been successful. 
\vspace{-0.01\textheight}
\begin{table}[h]
    \centering
    \caption{\small
    \textsc{:  Distance to Targets- Multi-Target MADDPG (first 2k episodes truncated)}}
        \begin{tabular}{|c|c|c|}
        \toprule
		\textbf{\textit{mean of min dist}} & \textbf{\textit{mean of ave dist}} & \textbf{\textit{mean of max dist}}\\ 
        \midrule
        $0.4616$  & $2.1048$ &  $3.8639$ \\
        \hline
        \end{tabular}
    \label{tab:Multi-tar dist}
\end{table}
\vspace{-0.01\textheight}
\subsection{Results for Role-based MADDPG Algorithm}
\begin{figure}[hb]
\centering
    \begin{subfigure}[b]{.7\linewidth}
        \centering
        \includegraphics[width=\linewidth]{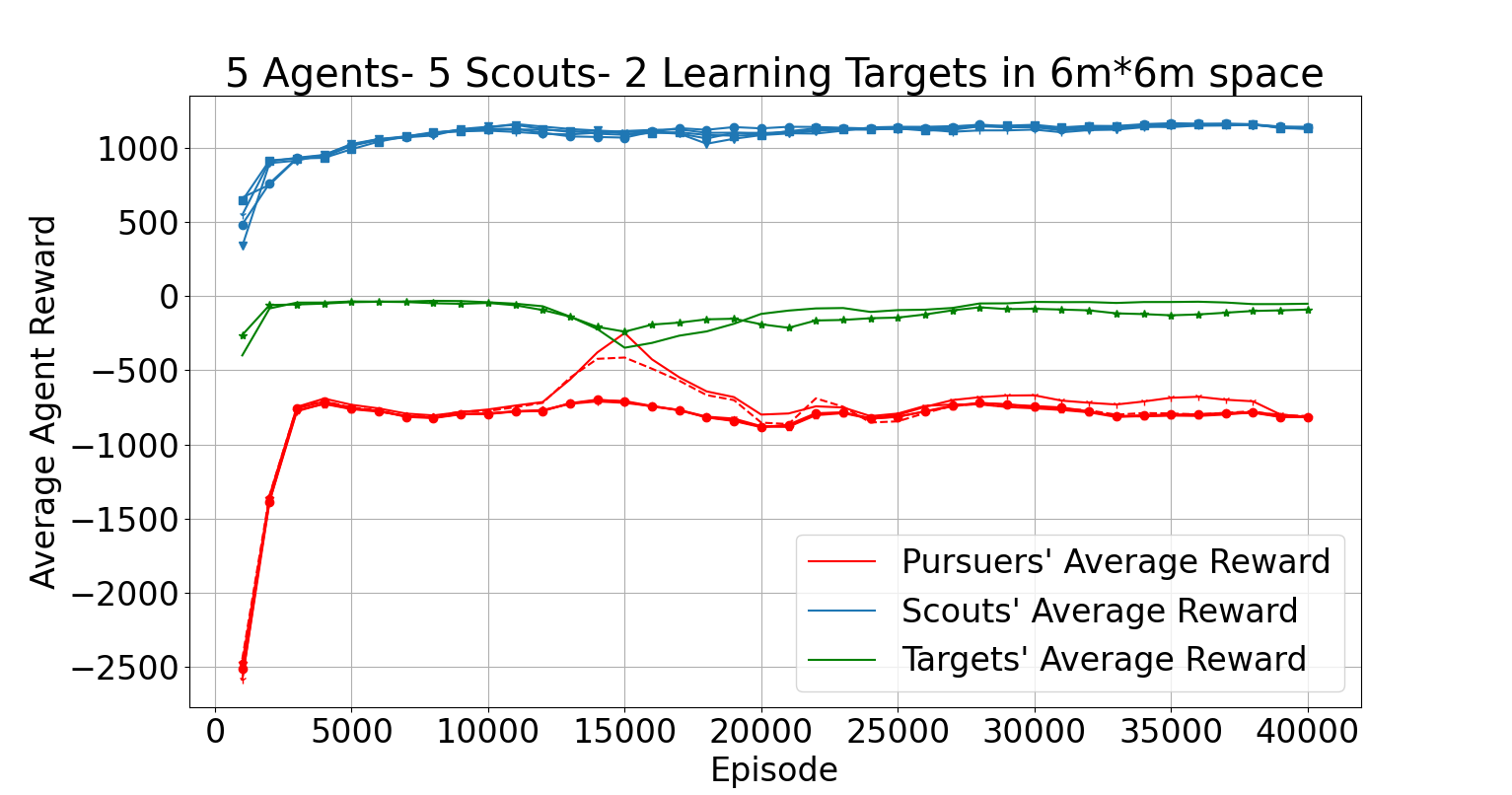}
        \caption{Average agent rewards} 
        \label{Sim:Role_based_rew}
    \end{subfigure}
    \begin{subfigure}[b]{.7\linewidth}
        \centering
        \includegraphics[width=\linewidth]{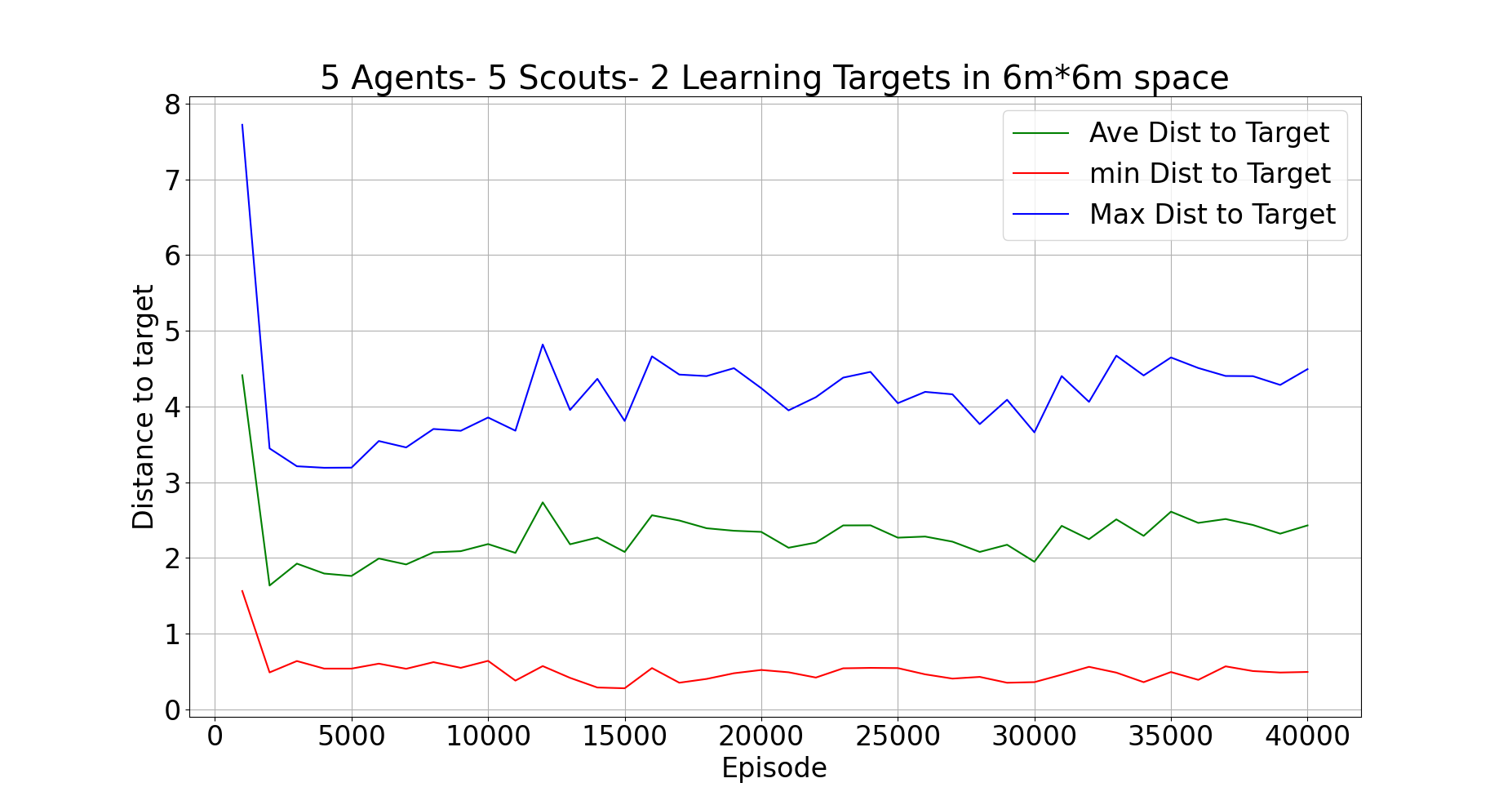}
        \caption{Distance to targets}
        \label{Sim:Role_based_dist}
    \end{subfigure}
\caption{Rewarding and distance-to-target metric for Multi-Target Role-based MADDPG (with 5 scouts and 5 pursuers vs. 2 evaders)}
\label{Sim:Multi-tar Role-based MADDPG}
\end{figure}

In the role-based MADDPG algorithm, we introduce scouts for environment exploration. In our results, evaders, pursuers and scouts are respectively trained with rewarding policies illustrated earlier in section~\ref{sec:Proposed approch}. 
Furthermore, the agents' average rewards resulting from training phase is shown in Fig.~\ref{Sim:Role_based_rew} where pursuers', evaders' and scouts' rewards are following their individual rewarding levels, and it is obvious that by adding scout agents, the reward level for pursuers and evaders is not effected due to the decentralized rewarding policies for heterogeneous agents. Meanwhile scouts and pursuers are interrelated via the communication network among neighbors (as mentioned in Section \ref{sec:ProbForm}, assumption (b)).
Finally Fig.~\ref{Sim:Role_based_dist} depicts the minimum, average and maximum distance to targets for Role-based MADDPG, and
a measure of distance to intelligent moving targets is summarized in Table.~\ref{tab:Role_based_dist}, and identically, the average minimum distance to the faster targets (observed as $0.4$~m) still can confirm a viable tracking and a successful pursuit mission. 
\vspace{-0.02\textheight}
\vspace{-0.01\textheight}
\begin{table}[h]
    \centering
    \caption{\small
    \textsc{:  Distance to Targets- Role-based MADDPG (first 2k episodes truncated)}}
        \begin{tabular}{|c|c|c|}
        \toprule
		\textbf{\textit{mean of min dist}} & \textbf{\textit{mean of ave dist}} & \textbf{\textit{mean of max dist}}\\ 
        \midrule
        $0.4745$  & $2.2609$ &  $4.1233$ \\
        \hline
        \end{tabular}
    \label{tab:Role_based_dist}
\end{table}
\vspace{-0.02\textheight}

\begin{figure}[htbp]
\centering
    \begin{subfigure}[b]{.49\linewidth}
        \centering
        \includegraphics[width=\linewidth,height=0.175\textheight]{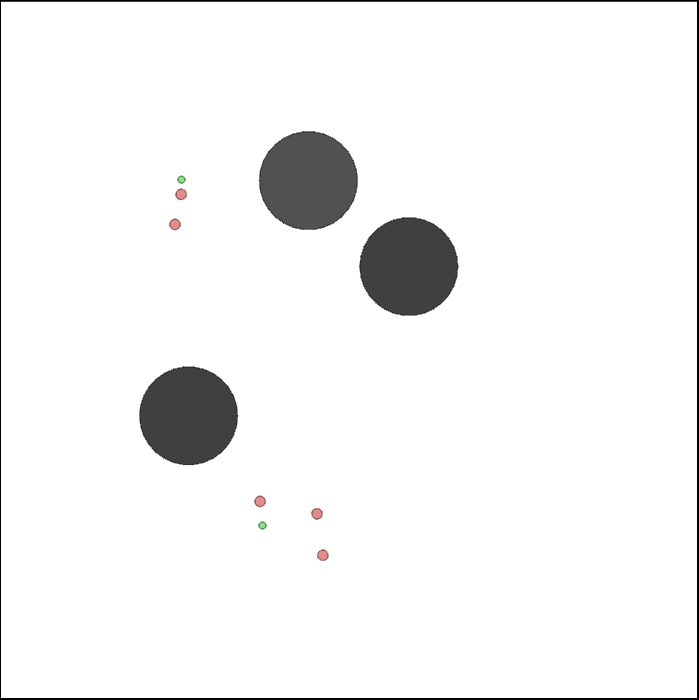}
        \caption{}
        \label{Sim:PE_result_screen}
    \end{subfigure}
    \begin{subfigure}[b]{.49\linewidth}
        \centering
        \includegraphics[width=\linewidth]{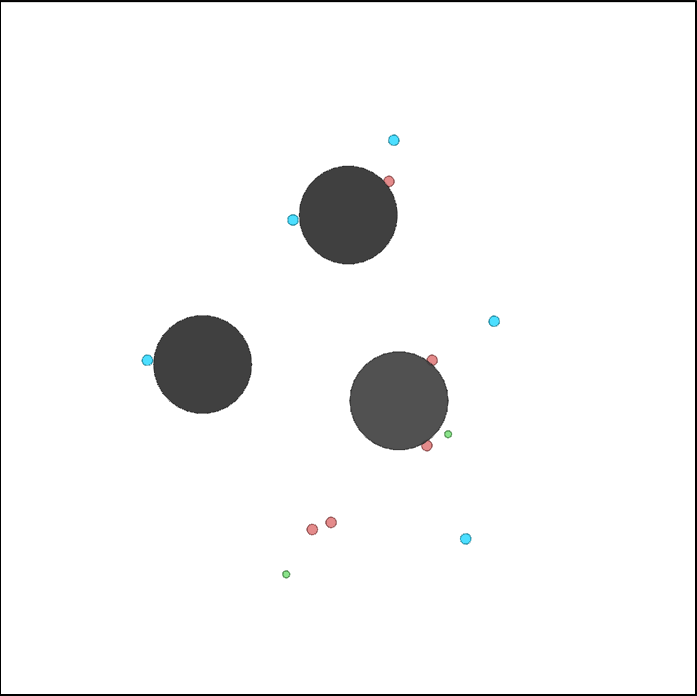}
        \caption{}
        \label{Sim:Role_based_pic}
    \end{subfigure}
\caption{Qualitative results of multi-target (a) MADDPG vs. (b) Role-based MADDPG with 2 targets (in green), 5 pursuers (in red) and 5 scouts (in blue) in $6$~m $\times$ $6$~m space}
\label{Sim:Qual_res}
\end{figure}

Lastly, compared to the original MADDPG algorithm, our proposed role-based MADDPG algorithm has not only been able to track multiple targets, but also has been able to explore the overlooked parts of space by means of the the proposed Voronoi-based rewarding policy. To serve a qualitative comparison, Fig.~\ref{Sim:PE_result_screen} shows the final result of multi-target MADDPG for one execution phase where 2 evading targets are shown in green, 5 pursuing agents are shown in red and randomized static obstacles represented by large circles, and in Fig.~\ref{Sim:Role_based_pic} a qualitative result of Role-based MADDPG is given incorporating 5 scouts (in blue) into the initial multi-target scenario, and we finally find the pursuers tracking the targets and the scouts exploring the environment (see the demo video: \text{https://youtu.be/xd6EaWNINB8}).

\section{Real-world results with Crazyflie Drones}
\label{sec:illExp}
\begin{figure}[htbp]
\centering
    \begin{subfigure}[h]{.25\textwidth}
        \centering
        \includegraphics[width=\textwidth]{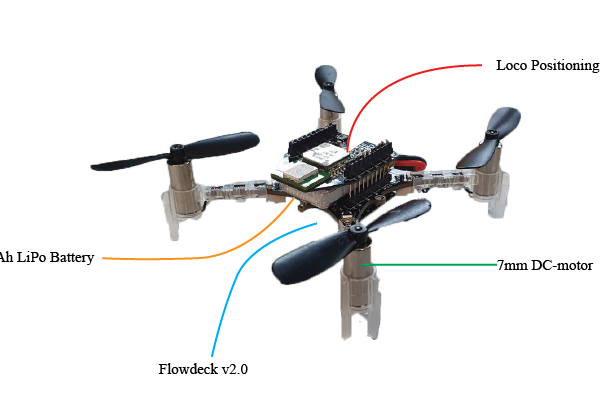}
        \caption{}
        \label{crazyflie_decks}
    \end{subfigure}
    \begin{subfigure}[h]{.85\linewidth}
        \centering
        \includegraphics[width=\linewidth]{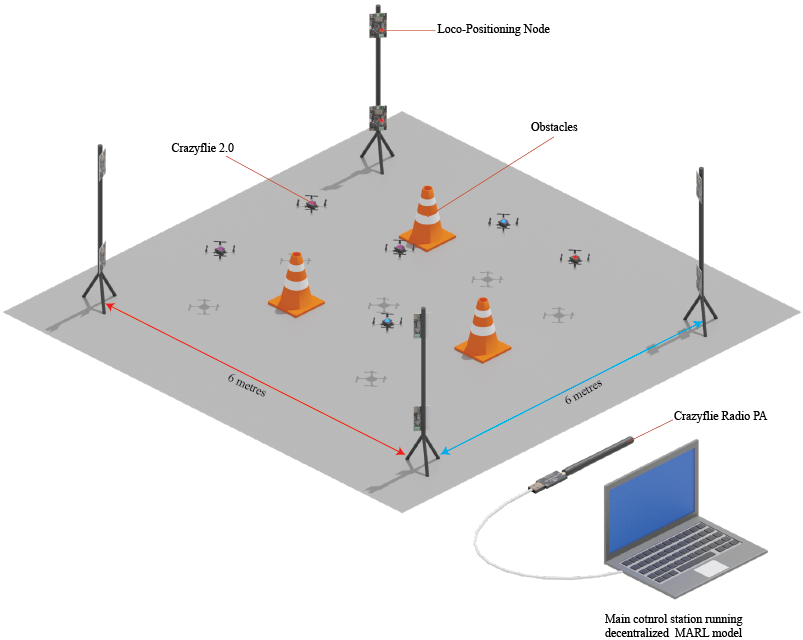}
        \caption{}
        \label{exp_setup}
    \end{subfigure}
\caption{Real experimental environment. (a) An illustration of Crazyflie decks. (b) An illustration of the experimental test-bed including the UWB anchors, drones, safety cones, and the radio antenna connected to compute.}
\label{Real_Exp}
\end{figure}

For the real-world experiments, we used \textit{Crazyflie 2.1} nanocopter developed by Bitcraze to showcase our role-based MARL framework. Given the small form factor of the drones, we used an external computer for computing needs. The drones communicate with the computer, using Bitcraze's \textit{Crazyradio PA}. Additionally, \textit{Loco-Positioning Decks} were attached to each drone along with \textit{Loco-Positioning Nodes} on each corner of the experimental area. The \textit{Loco-Positioning System} works based on Ultra Wide Band (UWB) technology and allows us to send position-based movement commands from the computer using the radio communication channels which can be set using Bitcraze's Crazyflie Client. In order to sense the height of flight for the drones, a \textit{Flowdeck v2} was attached to the bottom of each drone. The decks of a Crazyflie drone mentioned above, are illustrated in Fig.~\ref{crazyflie_decks}.
For performance assessment of above-mentioned MARL models, we set up an experimental environment of $6$~m $\times$ $6$~m that consists of the flight region, a computer as control station, a USB radio antenna and $4$ stands on the corners to hold UWB anchors for drones' positioning (two anchors attached to each stand to fulfill the drone localization with totally $8$ UWB anchors). Within the experimental environment depicted in Fig.~\ref{exp_setup}, we are using $3$ traffic safety cones as static obstacles and a total of $6$ drones ($3$ scouts, $2$ pursuers and $1$ faster evading target) have been utilized to validate our proposed Role-based MADDPG algorithm. 

\begin{figure}[htbp]
\centering
    \begin{subfigure}[b]{0.9\linewidth}
        \centering
        \includegraphics[width=\linewidth]{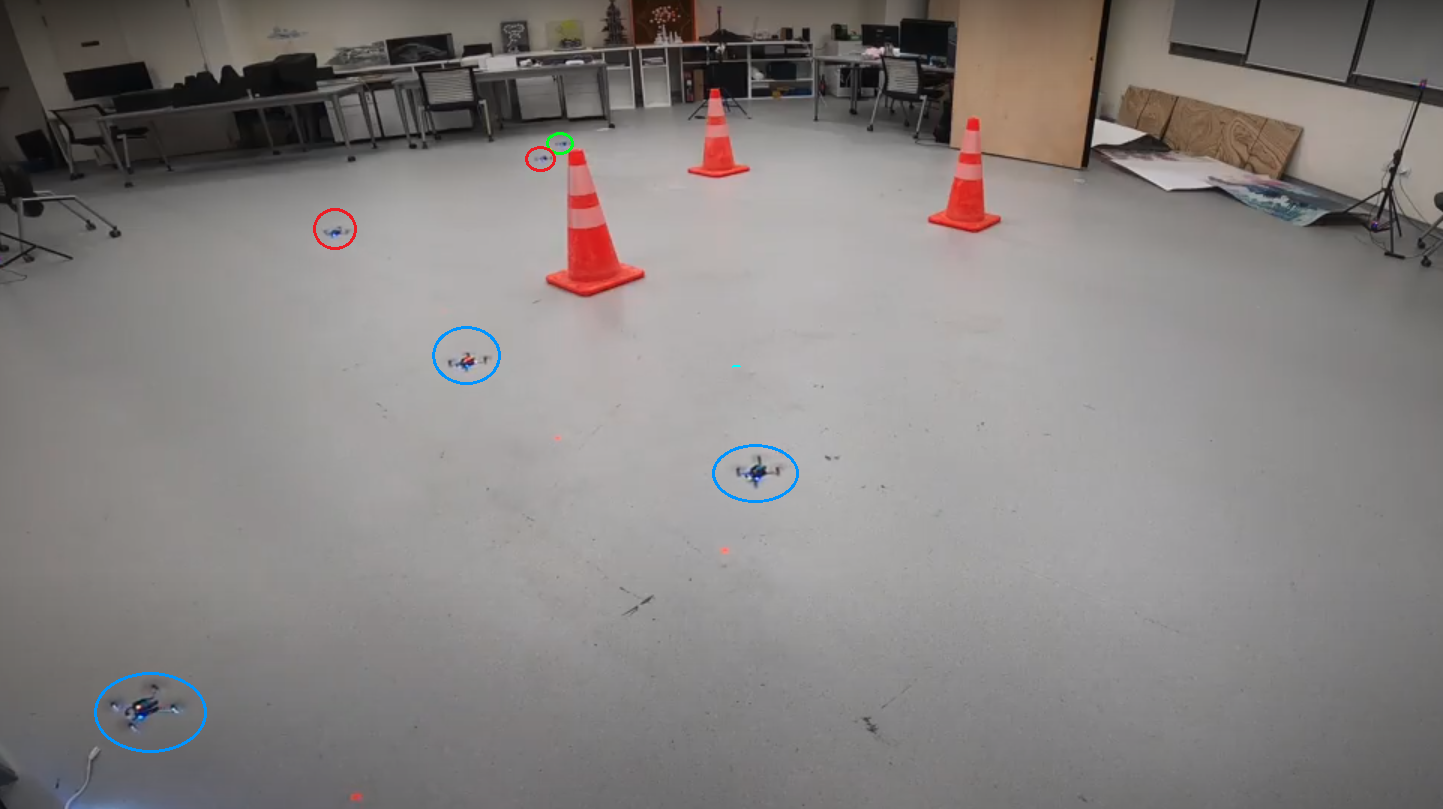}
        \caption{}
        \label{demo:Role_based_init}
    \end{subfigure}
    \begin{subfigure}[b]{0.9\linewidth}
        \centering
        \includegraphics[width=\linewidth]{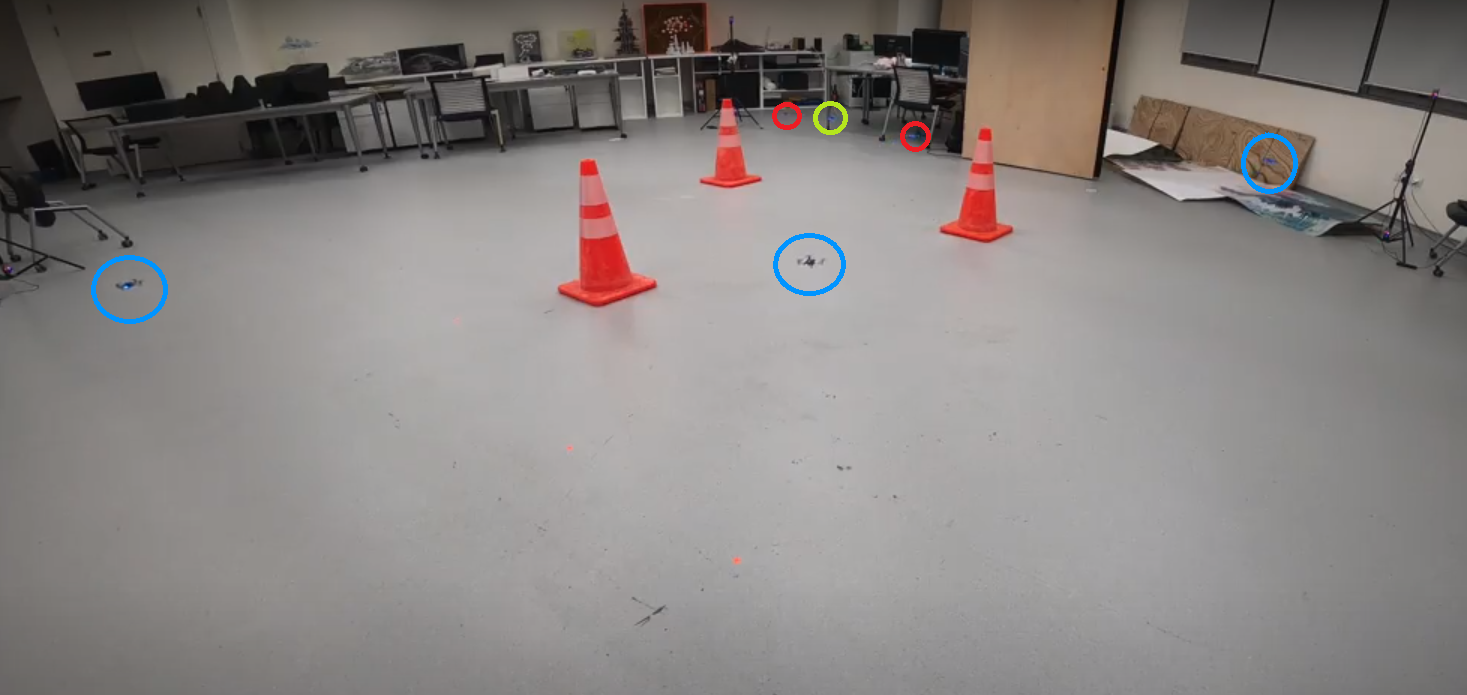}
        \caption{}
        \label{demo:Role_based_fin}
    \end{subfigure}
    \begin{subfigure}[b]{0.9\linewidth}
        \centering
        \includegraphics[width=\linewidth]{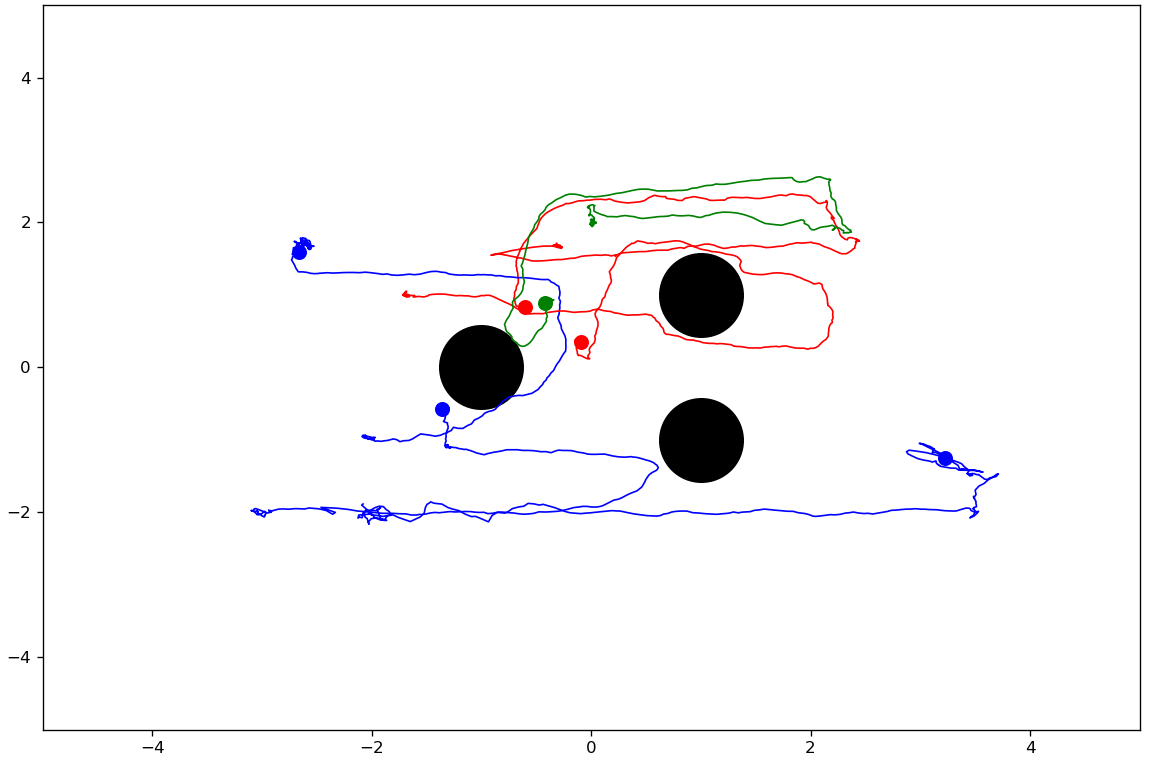}
        \caption{}
        \label{bird_view_trajectory}
    \end{subfigure}
\caption{Decentralized heterogeneous UAV swarm with Role-based MADDPG- drone demo on multi-player pursuit-evasion with 2 pursuers in red, 1 intelligent evader in green and 3 scouts in 
blue. (a) Initial frame, (b) Final frame, (c) Complete bird-eye-view trajectory from UWB readings.  }
\label{demo:Role_based}
\end{figure}

At the start of the experiment, we place 6 drones at their initial positions within the experimental environment and let the swarm fly through the region commanded by MARL model. The mission is considered complete when the pursuers, scouts and evaders are showing expected (learned) heterogeneous behaviors of target tracking, exploration and evasion respectively. 
Screenshot of the demo are illustrated in Fig.~\ref{demo:Role_based} where Fig.~\ref{demo:Role_based_init} shows the initial starting points of the agents and Fig.~\ref{demo:Role_based_fin} illustrates the final position of the agents after showcasing the learned behavior for corresponding heterogeneous roles. Colored annotations in the provided demo screenshots are following the agents' color code in OpenAI simulation results (red pursuers, green target and 
blue scouts). In Fig.~\ref{demo:Role_based_fin}, we finally observe that the pursuers are continuously keeping track of the target and the scouts are fairly scattered in the environment to handle exploration. In addition, Fig.~\ref{bird_view_trajectory} illustrates the full trajectory of UWB readings from real drone experiment where at the end of the pursuit mission, based on the experimental UWB data, the average distance to target (from pursuers) is calculated to be $0.4983$~m. This distance falls under the sensory range of the robot ($50$~cm) and the pursuit mission is perceived to be successful in this regard. Please see the demo video for further illustrations on the real drone experiment along with OpenAI simulation outputs and real-time UWB readings. 


\section{Conclusion and Future Works}\label{conc}
In this paper, based on multi-agent reinforcement learning (MARL), a collaborative/competitive framework is considered for a swarm of UAV agents which are tasked to fulfill pursuit-evasion and also carry out successful exploration. Collaborative task is performed among pursuers and scouts to fulfill the pursuit-exploration mission while pursuers are competing with evaders in terms of target tracking. We have proposed a variant of a MARL algorithm called Role-based Multi-Agent Deep Deterministic Policy Gradient (MADDPG) and apart from introducing rewarding policies for MARL framework to generalize towards a multi-target pursuit-evasion scenario, we have proposed  an approach based on Voronoi-cells concept to develop exploration on top of the multi-target pursuit-evasion. The MARL training and simulation results are implemented in MPE environment from OpenAI and the real world experiments are provided on Crazyflie drones. The simulation and real-world results show that the trained models have well adopted the expected behaviors and finally we find the pursuers successfully tracking the faster multiple intelligent targets and also we have some agents involved in exploration task. As a scope for future work, this study can be generalized towards specific applications by designing required swarm behaviors, and considering partial observability among the swarm to rely on emergent communication networks.  

\section*{Acknowledgement}
This research/project is supported by the National Research Foundation Singapore and DSO National Laboratories under the AI Singapore Programme (AISG Award No: AISG2-RP-2020-016).

\bibliographystyle{IEEEbib}
\bibliography{refs}

\end{document}